\documentclass[conference]{IEEEtran}

\makeatletter
\def\ps@IEEEtitlepagestyle{%
  \def\@oddfoot{\mycopyrightnotice}%
  \def\@oddhead{\hbox{}\@IEEEheaderstyle\leftmark\hfil}\relax
  \def\@evenhead{\@IEEEheaderstyle\hfil\leftmark\hbox{}}\relax
  \def\@evenfoot{}%
}
\def\mycopyrightnotice{%
  \begin{minipage}{\textwidth}
  \centering \scriptsize
  Copyright~\copyright~2025 IEEE. Personal use of this material is permitted. Permission from IEEE must be obtained for all other uses, in any current or future media, including reprinting/republishing this material for advertising or promotional purposes, creating new collective works, for resale or redistribution to servers or lists, or reuse of any copyrighted component of this work in other works.
  \end{minipage}
}

\makeatother

\IEEEoverridecommandlockouts

\usepackage{cite}
\usepackage{amsmath,amssymb,amsfonts}
\usepackage{algorithmic}
\usepackage{graphicx}
\usepackage{textcomp}
\usepackage{xcolor}
\usepackage{texnames}
\usepackage{siunitx}
\usepackage{booktabs}
\usepackage{multirow} 
\usepackage[colorlinks=true, urlcolor=blue, linkcolor=black]{hyperref}

\usepackage{float}

\begin{document}


\title{FuseTen: A Generative Model for Daily 10\,m Land Surface Temperature Estimation from Spatio-Temporal Satellite Observations}


\author{\IEEEauthorblockN{Sofiane Bouaziz\IEEEauthorrefmark{1}\IEEEauthorrefmark{2}, Adel Hafiane\IEEEauthorrefmark{1}, Raphaël Canals \IEEEauthorrefmark{2}, Rachid Nedjai \IEEEauthorrefmark{3}}

\IEEEauthorblockA{\IEEEauthorrefmark{1} INSA CVL, Université d’Orléans, PRISME UR 4229, Bourges, 18022, Centre Val de Loire, France }

\IEEEauthorblockA{\IEEEauthorrefmark{2}Université d’Orléans, INSA CVL, PRISME UR 4229, Orléans, 45067, Centre Val de Loire, France}

\IEEEauthorblockA{\IEEEauthorrefmark{3} Université d'Orléans, CEDETE, UR 1210, Orléans, 45067, Centre Val de Loire, France}}

\maketitle

\begin{abstract}
Urban heatwaves, droughts, and land degradation are pressing and growing challenges in the context of climate change. A valuable approach to studying them requires accurate spatio-temporal information on land surface conditions. One of the most important variables for assessing and understanding these phenomena is Land Surface Temperature (LST), which is derived from satellites and provides essential information about the thermal state of the Earth's surface. However, satellite platforms inherently face a trade-off between spatial and temporal resolutions.  To bridge this gap, we propose \textit{FuseTen}, a novel generative framework that produces daily LST observations at a fine 10 m spatial resolution by fusing spatio-temporal observations derived from Sentinel-2, Landsat 8, and Terra MODIS. 
\textit{FuseTen} employs a generative architecture trained using an averaging-based supervision strategy grounded in physical principles. It incorporates attention and normalization modules within the fusion process and uses a PatchGAN discriminator to enforce realism. Experiments across multiple dates show that \textit{FuseTen} outperforms linear baselines, with an average 32.06\% improvement in quantitative metrics and 31.42\% in visual fidelity. To the best of our knowledge, this is the first non-linear method to generate daily LST estimates at such fine spatial resolution. 

\end{abstract}


\section{Introduction}
\label{sec:introduction}
\IEEEPARstart{A}{s} climate change accelerates and urban areas expand, understanding the dynamics of land surface behavior has become increasingly important~\cite{roy2022anthropogenic}. Land Surface Temperature (LST) is a key variable for analyzing and interpreting diverse environmental phenomena. It represents the radiative temperature emitted by the Earth’s surface, influenced by its physical and biological properties~\cite{HULLEY201957}. LST plays a crucial role in diverse applications such as climate change assessment, agricultural monitoring, and urban heat island studies~\cite{almeida2021study}. Recognizing its significance in land–atmosphere interactions, the Global Climate Observing System has designated LST as one of the ten essential climate variables~\cite{hollmann2013esa}.


\smallskip

\noindent Remote sensing (RS) satellites remain the primary approach for monitoring LST~\cite{li2013satellite}, yet it is constrained by a trade-off between spatial and temporal resolutions. Spatial resolution defines the level of detail captured within each LST pixel~\cite{ZHANG2023129}, while temporal resolution refers to how frequently LST measurements are acquired~\cite{20201Shunlin}. Achieving high spatial and temporal resolution simultaneously is crucial.



\smallskip

\noindent Spatio-Temporal Fusion (STF) techniques provide an alternative approach to generate observations with both high spatial and temporal resolutions by combining RS satellites that differ in these characteristics~\cite{bouaziz2024deep}. Classical STF methods rely on physical or statistical models that assume linear relationships between spatial and temporal variations. This linearity limits their ability to model complex interactions in heterogeneous landscapes, leading to reduced accuracy~\cite{zhan2016disaggregation}.

\smallskip

\noindent Recent advances in deep learning (DL) have demonstrated strong capabilities in capturing complex, non-linear spatio-temporal patterns~\cite{bouaziz2024deep}. Specifically, Generative Adversarial Networks (GANs)~\cite{goodfellow2014generative} have been applied to STF, where the generator produces high-resolution fused images and the discriminator assesses their realism~\cite{song2022mlff, huang2024stfdiff}. However, very few studies have applied DL specifically to the STF for LST estimation. Moreover, to the best of our knowledge, no prior work has successfully produced daily LST estimates at very high spatial resolution of 10 m using a DL model. In this study, we propose a novel deep generative framework for daily 10 m LST estimation through STF of Sentinel-2, Landsat 8, and Terra MODIS data. Our key contributions are as follows:


\begin{itemize}
    \item We present the first non-linear generative model specifically designed for STF of LST, allowing daily temperature estimation at 10 m resolution using only 1 km Terra MODIS LST and auxiliary spectral information.
    \item We introduce an innovative, physically motivated supervision strategy based on 30 m Landsat-derived LST  to bypass the lack of ground truth at 10 m resolution.
    \item We demonstrate the effectiveness of our approach on a real-world dataset and show its superiority.
\end{itemize}

\vspace{-0.25em}

\section{Background}
\label{sec:Related_works}
\begin{figure*}[htbp]
  \centering
  \includegraphics[width=0.9\textwidth]{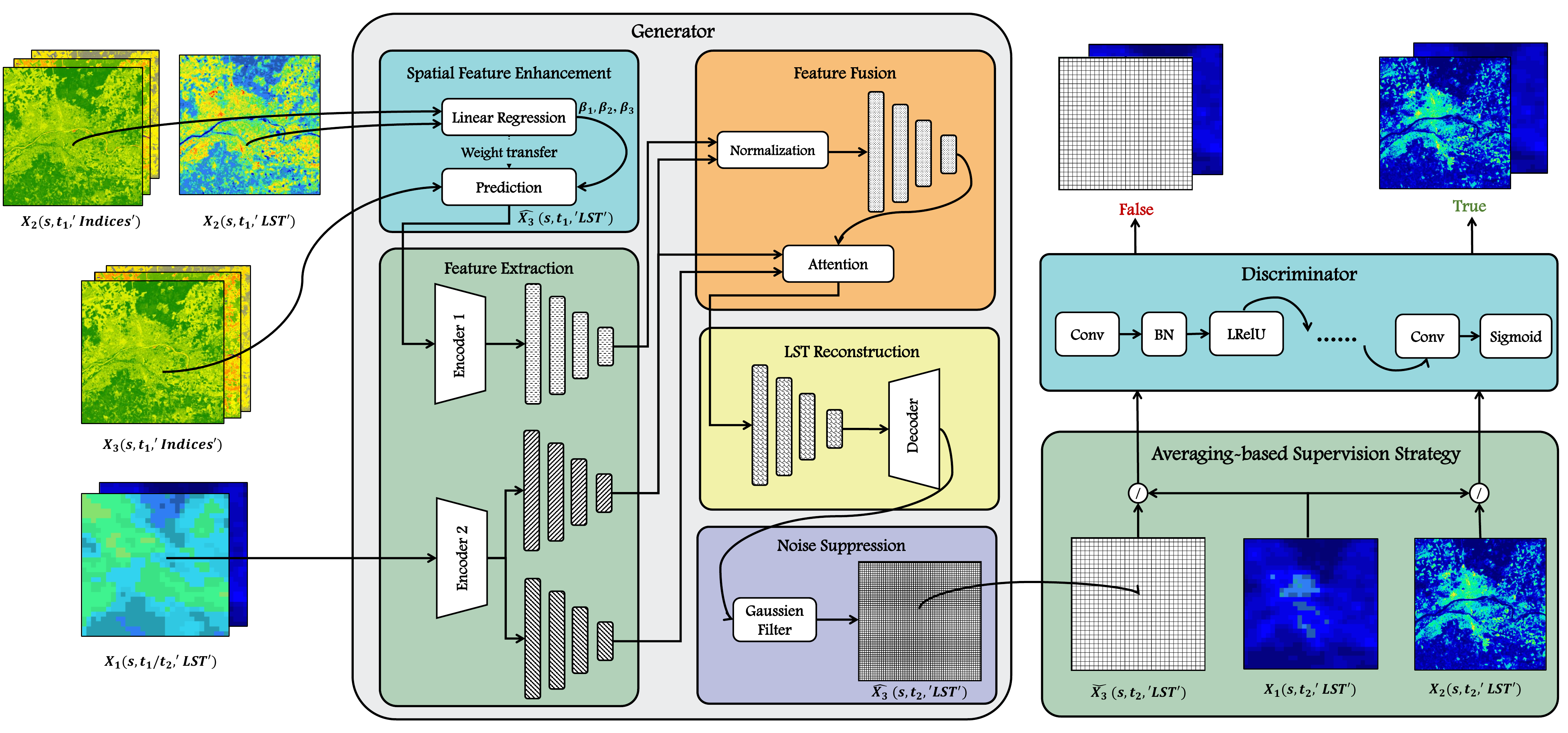}
  \caption{ \textit{FuseTen} architecture. The generator fuses multi-resolution satellite inputs to estimate 10\,m LST at the target date $t_2$. It leverages spectral indices, Terra MODIS LST, and an averaging-based supervision strategy guided by a PatchGAN discriminator.}
  \label{fig:architecture}
\end{figure*}

\subsection{Satellite-based LST}
LST refers to the radiometric temperature of the Earth’s surface~\cite{norman1995terminology}. It is primarily retrieved from RS satellite using thermal infrared sensors. However, its accuracy is limited by the trade-off between spatial and temporal resolution.



       

\noindent Table~\ref{tab:lstsat} summarizes representative satellites that illustrates this trade-off. Terra MODIS offers daily LST at 1 km resolution, while Landsat 8 provides finer 30 m LST with a 16-day revisit cycle. Sentinel-2, despite lacking thermal sensors, still contributes high-resolution optical data (10 m, 5-day revisit).

\vspace{-1em}

\begin{table}[ht]
\centering
\caption{Comparison of representative satellite products with Their Thermal Sensors, Spatial and Temporal Resolutions.}
\vspace{-1.25em}
\renewcommand{\arraystretch}{1} 
\begin{tabular}[t]{@{\hspace{2pt}}l@{\hspace{6pt}}c@{\hspace{6pt}}c@{\hspace{6pt}}c@{\hspace{2pt}}}
\hline
\textbf{Satellite} & \textbf{Thermal Sensor} & \textbf{Spatial Resolution} & \textbf{Temporal Resolution} \\

\hline

Terra & MODIS & 1 km & 1 day  \\

Landsat 8 & TIRS & 30 m & 16 days \\

Sentinel-2 & / & 10 m & 5 days \\

\hline
\label{tab:lstsat}
\end{tabular}
\end{table}

\vspace{-1em}

\subsection{Spatio-Temporal Fusion for LST estimation}
\label{sec:STF}

STF addresses the trade-off between spatial and temporal resolution in RS satellite by framing it as a multi-objective optimization problem that aims to simultaneously enhance both resolutions~\cite{bouaziz2024deep}.  Let \( X_1 \) and \( X_2 \) denote LST observations from low-spatial/high-temporal (LSHT) and high-spatial/low-temporal (HSLT) sensors, respectively. For a geographical region \( s \) and two time steps \( t_1 \) and \( t_2 \), we define the data pair at \( t_1 \) as \( P_1 = \{X_1(s, t_1, \text{`LST'}),\; X_2(s, t_1, \text{`LST'})\} \). Given \( P_1 \) and the LSHT observation \( X_1(s, t_2, \text{`LST'}) \), the goal is to estimate the missing HSLT observation at $t_2$, \( \hat{X}_2(s, t_2, \text{`LST'}) \). This fusion is modeled by a DL network \( f(\cdot \mid \mathbf{W}) \), where \( \mathbf{W} \) are trainable parameters, as shown in Equation~\ref{eq:stf_dl}.




\vspace{-1em}

\begin{equation}
\hat{X}_2(s, t_2, \text{`LST'}) = f\left(P_1,\; X_1(s, t_2, \text{`LST'}) \mid \mathbf{W} \right).
\label{eq:stf_dl}
\end{equation}

\section{Proposed Approach}

\textit{FuseTen} redefines the STF problem outlined in Section \ref{sec:STF} by leveraging three complementary satellite sources, each with distinct spatial and temporal resolutions: Terra MODIS (\( X_1 \)), Landsat 8 (\( X_2 \)), and Sentinel-2 (\( X_3 \)).  The overall architecture, shown in Figure~\ref{fig:architecture}, is based on a Conditional GAN, where the generation process is conditioned on Terra MODIS LST observation at the target date. The goal is to predict the LST at 10\,m for a given geographical region \( s \) and a target time \( t_2 \). The final output of the model is denoted as \( \hat{X}_3(s, t_2, \text{`LST'}) \). To perform this prediction, we first collect the MODIS LST at \( t_2 \), denoted as \( X_1(s, t_2, \text{`LST'}) \), available at 1 km resolution. Next, we identify a prior reference date \( t_1 \) with minimal cloud cover, where Terra MODIS, Landsat 8, and Sentinel-2 all have overlapping observations over the same region. From this reference date, $t_1$, we extract the following Triple $T_1$ :


\smallskip


\begin{itemize}
    \item Sentinel-2 spectral indices at 10\,m:
    \( X_3(s, t_1, \text{`NDVI'}) \), \( X_3(s, t_1, \text{`NDWI'}) \), \( X_3(s, t_1, \text{`NDBI'}) \).
    \item Landsat 8 spectral indices at 30\,m:
    \( X_2(s, t_1, \text{`NDVI'}) \), \( X_2(s, t_1, \text{`NDWI'}) \), \( X_2(s, t_1, \text{`NDBI'}) \).
    \item Landsat 8 LST at 30\,m:
    \( X_2(s, t_1, \text{`LST'}) \).
    \item Terra MODIS LST at 1\,km:
    \( X_1(s, t_1, \text{`LST'}) \).
\end{itemize}

\smallskip

\noindent These inputs are fed to the generator, which learns to extract spatio-temporal and spatio-spectral features to produce a 10 m LST map at \( t_2 \), \( \hat{X}_3(s, t_2, \text{`LST'}) \). Without ground truth at 10 m, we adopt an averaging-based supervision strategy, where the generated LST is averaged using a 3\,$\times$\,3 window to approximate the 30 m, matched against Landsat 8 LST, and both are passed to the discriminator for adversarial training. The framework’s components are detailed in the next subsections.


\subsection{Generator}

The generator \( G \) is structured into five key stages: \textit{spatial feature enhancement}, \textit{feature extraction}, \textit{feature fusion}, \textit{image reconstruction}, and \textit{Noise suppression}.

\noindent In the \textit{spatial feature enhancement} stage, we generate a preliminary 10 m LST for the reference date \( t_1 \). To do this, a linear regression model is trained using Landsat 8 spectral indices (NDVI, NDWI, NDBI) at 30 m to predict Landsat-derived LST, learning weights \( \beta_1, \beta_2, \beta_3 \), as described in Equation \ref{eq:lr}. These learned weights are then applied to Sentinel-2 spectral indices to produce a 10 m LST estimate at \( t_1 \), \( \hat{X_3}(s, t_1, \text{`LST'}) \).

\begin{equation}
\begin{aligned}
\hat{X}_2(s, t_1, \text{'\textit{LST}'}) &= \beta_1 X_2(s, t_1, \text{'\textit{NDVI}'}) + \beta_2 X_2(s, t_1, \text{'\textit{NDWI}'}) \\
&\quad + \beta_3 X_2(s, t_1, \text{'\textit{NDBI}'})
\end{aligned}
\label{eq:lr}
\end{equation}

\noindent The \textit{feature extraction} stage uses several convolutional blocks with progressive downsampling to encode spatial and temporal features into a latent representation. Residual blocks are used to retain fine spatial details while omitting batch normalization. \noindent Next, the \textit{feature fusion} module performs two key operations. First, \textit{Adaptive Instance Normalization}~\cite{huang2017arbitrary} aligns the statistical distributions of features originating from different temporal sources by harmonizing the spatial characteristics captured by Sentinel-2 and Landsat 8 with the temporal patterns present in Terra MODIS. Second, a \textit{temporal attention} mechanism adaptively weights feature importance across time, which allows the model to prioritize relevant temporal information. The \textit{image reconstruction} stage mirrors the encoder in a U-Net-like architecture, combining upsampling, transposed convolutions, and residual blocks to decode the fused features and generate the final 10 m LST estimation at the target date, \( \hat{X}_3(s, t_2, \text{`LST'}) \). Finally, a \textit{noise suppression} stage is applied using a Gaussian filter (\( \mu = 0, \sigma = 1 \)) to reduce high-frequency artifacts and noise introduced by convolutions. 

\subsection{Averaging-based Supervised Learning}

Since no ground-truth LST is available at 10 m, we introduce an averaging-based, physically motivated supervision strategy based on the assumption that coarse-resolution LST can be approximated by the local averaging of fine-resolution values~\cite{gao2016localization}. Specifically, a \(3 \times 3\) average pooling is applied to the generator’s 10 m output to approximate 30 m resolution, which is then compared to Landsat 8 LST \( X_2(s, t_2, \text{‘LST’}) \).




\subsection{Discriminator}

The discriminator adopts a PatchGAN architecture~\cite{goodfellow2014generative}. Unlike conventional GANs that distinguish real from fake images, our discriminator is trained to identify whether a given LST image is a real observation or a fused result. It is conditioned on Terra MODIS LST at time \( t_2 \), \( X_1(s, t_2, \text{‘LST’}) \), which provides reliable 1 km resolution thermal context. This conditioning helps the discriminator better interpret the thermal scene and assess the plausibility of the input LST map. A sigmoid layer at the output is employed to produce probability scores. During training, the discriminator is encouraged to output values close to 1 when presented with the observed Landsat 8 LST \( X_2(s, t_2, \text{‘LST’}) \), and values near 0 when evaluating the averaged fused LST estimate \( \hat{X}_3(s, t_2, \text{‘LST’}) \), alongside the corresponding Terra MODIS reference \( X_1(s, t_2, \text{‘LST’}) \).


\section{Evaluation Methodology and Results}
\subsection{Region of Interest}

\begin{figure}[htbp]
  \centering
  \includegraphics[width=0.5\textwidth]
  {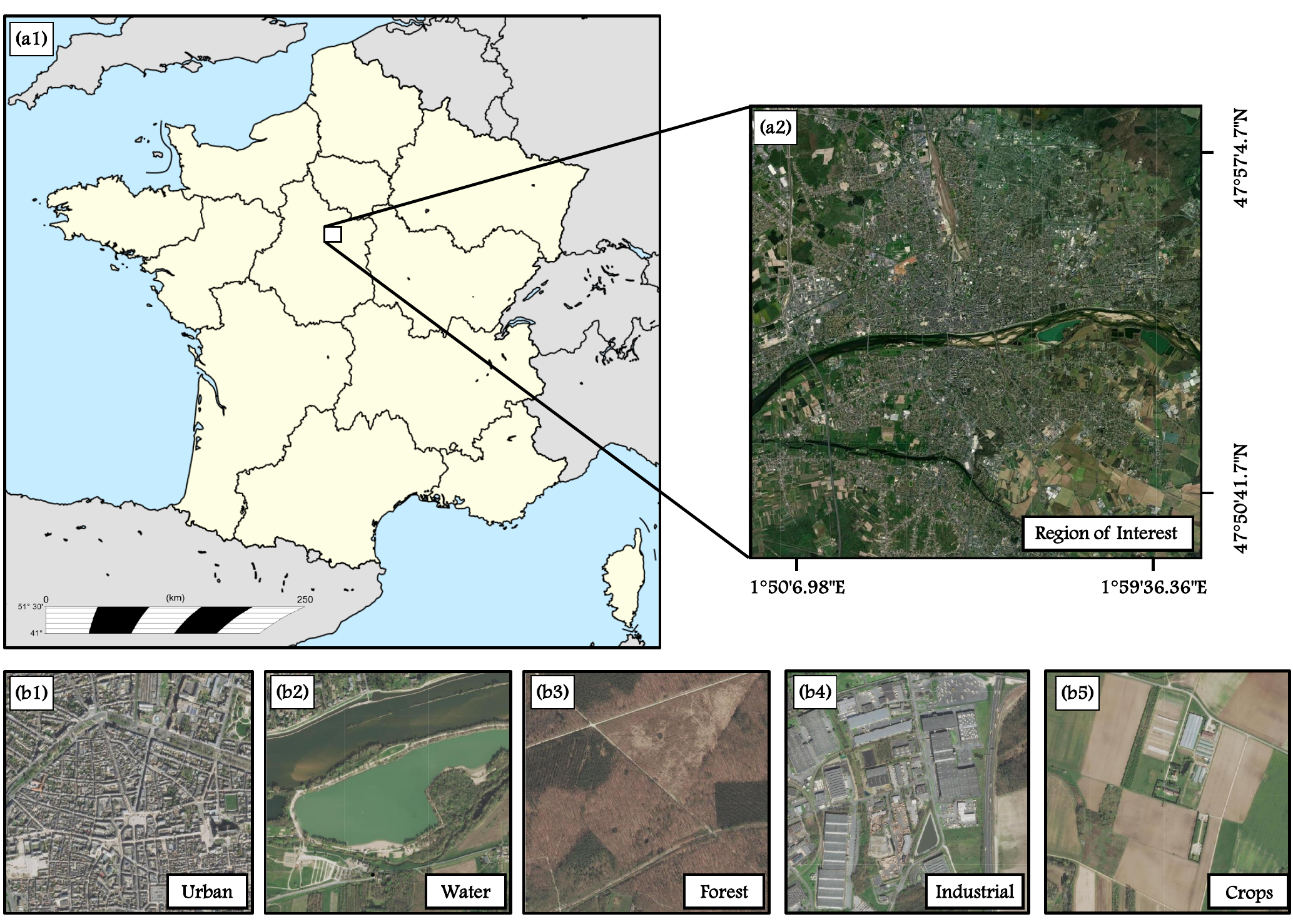}
  \vspace{-1em}
    \caption{Overview of the ROI. (a1) Location of the ROI in France. (a2) Satellite view of the selected ROI, (b1)–(b5) Examples of dominant land cover types within the ROI: urban, water, forest, industrial, and agricultural areas.}
    
  \label{fig:ROI}
\end{figure}

The region of interest is located within Orléans Métropole in the Centre-Val de Loire region of France. As shown in Fig.\ref{fig:ROI}, it spans approximately 144\,km$^2$, between latitudes 47$^\circ$50'41.77''N–47$^\circ$54'1.74''N and longitudes 1$^\circ$50'6.98''E–1$^\circ$59'36.36''E. The Loire River, France’s longest, runs through the area, influencing both its geography and thermal dynamics. The ROI includes diverse land cover types including urban areas, water bodies, forests, industrial zones, and crops fields. This makes the ROI a suitable testbed for assessing the generalizability and robustness of \textit{FuseTen}.

\vspace{-0.8em}

\vspace{0.8em}

\begin{table*}[htbp]
\centering
\caption{Quantitative comparison of FuseTen against BicubicI and Ten-ST-GEE~\cite{mhawej2022daily} over three dates.}
\vspace{-0.5em}
\renewcommand{\arraystretch}{1.05} 
\begin{tabular}{@{\hskip 5pt}c@{\hskip 5pt} @{\hskip 5pt}c@{\hskip 5pt} @{\hskip 5pt}c@{\hskip 5pt} @{\hskip 5pt}c@{\hskip 5pt} @{\hskip 5pt}c@{\hskip 5pt} @{\hskip 5pt}c@{\hskip 5pt} @{\hskip 5pt}c@{\hskip 5pt} @{\hskip 5pt}c@{\hskip 5pt} @{\hskip 5pt}c@{\hskip 5pt} @{\hskip 5pt}c@{\hskip 5pt} @{\hskip 5pt}c@{\hskip 5pt} @{\hskip 5pt}c@{\hskip 5pt} @{\hskip 5pt}c@{\hskip 5pt}}
\hline
\textbf{Metric} & \textbf{Date} & \textbf{BicubicI} & \textbf{Ten-ST-GEE} & \textbf{FuseTen} & \textbf{Date} & \textbf{BicubicI}  & \textbf{Ten-ST-GEE} & \textbf{FuseTen} & \textbf{Date} & \textbf{BicubicI}  & \textbf{Ten-ST-GEE} & \textbf{FuseTen} \\
\hline
RMSE & \multirow{6}{*}{\rotatebox{90}{19-09-2024}} & 3.637 & 3.934 & \textbf{3.220} & 
\multirow{6}{*}{\rotatebox{90}{05-10-2024}} & 2.451 & 2.583 & \textbf{1.050} & 
\multirow{6}{*}{\rotatebox{90}{21-10-2024}} & 2.150 & 2.342 & \textbf{1.905} \\

SSIM & & 0.640 & 0.526 & \textbf{0.798} &
& 0.666 & 0.621 & \textbf{0.863} & 
& 0.770 & 0.837 & \textbf{0.867} \\

PSNR & & 15.357 & 14.503 & \textbf{18.552} & 
& 17.765 & 17.310 & \textbf{25.657} &
& 18.450 & \textbf{24.592} & 21.748  \\

SAM & & 4.841 & 6.168 & \textbf{3.865} & 
& 3.162 & 6.158 & \textbf{2.823}  &
& 3.930 & 4.689 & \textbf{3.033} \\

CC & & 0.572 & -0.025 & \textbf{0.814} &
& 0.571 & 0.113 & \textbf{0.901} & 
& 0.494 & 0.120 & \textbf{0.896} \\

ERGAS & & 4.595 & 4.983 & \textbf{3.944} & 
& 4.190 & 4.447 & \textbf{1.795} & 
& 3.315 & 3.610 & \textbf{2.937} \\

\hline
\label{tab:quantitive}
\end{tabular}
\end{table*}
\vspace{-2em}

\subsection{Data Preparation and Sampling Strategy}
We selected 20 image triplets where Sentinel-2, Landsat-8, and Terra MODIS observations overlap temporally and spatially with up to 80\% cloud cover. From these, 10 temporally non-overlapping samples corresponding to time steps $t_1$ and $t_2$ were created, with 7 samples used for training and 3 for testing. Each LST observation has a resolution of $1200 \times 1200$  pixels (Sentinel-2) and $400 \times 400$ pixels (Landsat-8). Missing values were filled using an adaptive spatial interpolation, which progressively expands a $3 \times 3$ window until valid neighbors were found. LST observations were then divided into $96 \times 96$ patches with a stride of 24, yielding 15,463 training samples. The model was trained with a learning rate of $2 \times 10^{-4}$ and a batch size of 32.


\subsection{Quantitative Assessment}

Table~\ref{tab:quantitive} compares the proposed \textit{FuseTen} method with two baselines. Since no DL method specifically targets this problem and most linear approaches lack public implementations, we selected \textit{BicubicI}, a bicubic interpolation of Terra MODIS LST to Sentinel-2 resolution, and \textit{Ten-ST-GEE}~\cite{mhawej2022daily}, a reference method that fuses Terra MODIS and Sentinel-2 via robust least squares. The test consists of predicting the LST at dates 19-09-2024, 2024-10-05, and 2024-10-21, using the Terra MODIS LST at the target date and a previous triple $T$ at an earlier date. To compute the performance, the predicted 10 m LSTs were averaged over a \(3 \times 3\) window to match 30 m resolution and validated against Landsat 8 LST at the target date. Evaluation is conducted over three dates using six metrics: RMSE, SSIM, PSNR, SAM, CC, and ERGAS. \textit{FuseTen} consistently outperforms both baselines. It achieves the lowest RMSE, SAM, and ERGAS, indicating superior accuracy, while attaining the highest SSIM, PSNR, and CC, demonstrating improved structural preservation, image quality, and correlation with reference LST. Notably, \textit{FuseTen} achieves an average RMSE reduction of 32.06\% and an average SSIM improvement of 31.42\% compared to \textit{Ten-ST-GEE}.


\subsection{Qualitative Assessment}

\begin{figure}[htbp]
  \centering
  \includegraphics[width=0.45\textwidth]{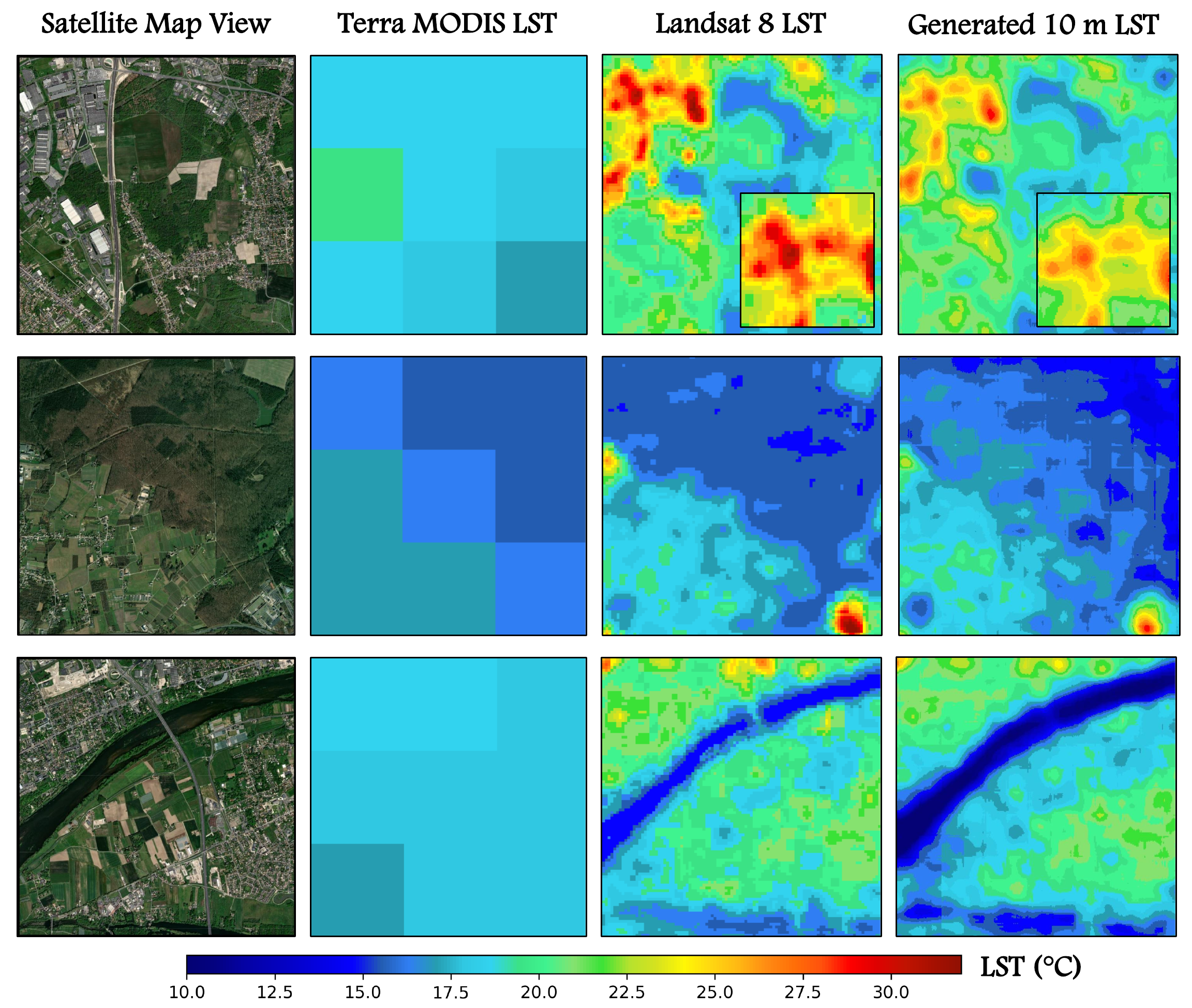}
    \caption{Qualitative comparison of LST maps from Terra MODIS, Landsat 8, and the proposed FuseTen method, alongside a high-resolution satellite view.}
    
  \label{fig:qualitative}
\end{figure}

Figure~\ref{fig:qualitative} presents a qualitative comparison between the Terra MODIS LST, reference Landsat 8 LST, the generated 10 m LST by FuseTen, and the corresponding high-resolution satellite map. We can observe that although Landsat 8 provides finer spatial resolution than Terra MODIS, its LST produces often lack visual consistency with actual land surface features. In contrast, the 10 m LST maps generated by FuseTen exhibit temperature distributions that closely align with real-world structures seen in the satellite imagery such as rivers and urban zones. This highlights FuseTen’s ability to enhance spatial detail while maintaining structural coherence. Additionally, unlike Landsat’s 16-day revisit cycle, FuseTen delivers these high-resolution LST on a daily basis. 

\vspace{-0.5em}

\vspace{-0.5em}

\section{Conclusion}
In this study, we introduced \textit{FuseTen}, a DL framework for fusing multi-resolution satellite observations to generate daily 10 m LST. \textit{FuseTen} leverages conditional GAN and an averaging-based learning strategy to enhance spatial and temporal resolutions. It reduces quantitative errors by an average of 32.06\% and improves visual fidelity by an average of 31.42\% compared to baseline methods.

\bibliographystyle{IEEEtranS}
\bibliography{MyReferences}

\begin{thebibliography}{10}
\providecommand{\url}[1]{#1}
\csname url@samestyle\endcsname
\providecommand{\newblock}{\relax}
\providecommand{\bibinfo}[2]{#2}
\providecommand{\BIBentrySTDinterwordspacing}{\spaceskip=0pt\relax}
\providecommand{\BIBentryALTinterwordstretchfactor}{4}
\providecommand{\BIBentryALTinterwordspacing}{\spaceskip=\fontdimen2\font plus
\BIBentryALTinterwordstretchfactor\fontdimen3\font minus \fontdimen4\font\relax}
\providecommand{\BIBforeignlanguage}[2]{{%
\expandafter\ifx\csname l@#1\endcsname\relax
\typeout{** WARNING: IEEEtranS.bst: No hyphenation pattern has been}%
\typeout{** loaded for the language `#1'. Using the pattern for}%
\typeout{** the default language instead.}%
\else
\language=\csname l@#1\endcsname
\fi
#2}}
\providecommand{\BIBdecl}{\relax}
\BIBdecl

\bibitem{almeida2021study}
C.~R.~d. Almeida, A.~C. Teodoro, and A.~Gon{\c{c}}alves, ``Study of the urban heat island (uhi) using remote sensing data/techniques: A systematic review,'' \emph{Environments}, vol.~8, no.~10, p. 105, 2021.

\bibitem{bouaziz2024deep}
S.~Bouaziz, A.~Hafiane, R.~Canals, and R.~Nedjai, ``Deep learning for spatio-temporal fusion in land surface temperature estimation: A comprehensive survey, experimental analysis, and future trends,'' \emph{arXiv preprint arXiv:2412.16631}, 2024.

\bibitem{gao2016localization}
L.~Gao \emph{et~al.}, ``Localization or globalization? determination of the optimal regression window for disaggregation of land surface temperature,'' \emph{IEEE Transactions on Geoscience and Remote Sensing}, vol.~55, no.~1, pp. 477--490, 2016.

\bibitem{20201Shunlin}
P.~Gibson, ``Chapter 1 - a systematic view of remote sensing (second edition),'' in \emph{Advanced Remote Sensing}, S.~Liang and J.~Wang, Eds.\hskip 1em plus 0.5em minus 0.4em\relax Academic Press, 2020, pp. 1--57.

\bibitem{goodfellow2014generative}
I.~Goodfellow \emph{et~al.}, ``Generative adversarial nets,'' \emph{Advances in neural information processing systems}, vol.~27, 2014.

\bibitem{hollmann2013esa}
R.~Hollmann \emph{et~al.}, ``The esa climate change initiative: Satellite data records for essential climate variables,'' \emph{Bulletin of the American Meteorological Society}, vol.~94, no.~10, pp. 1541--1552, 2013.

\bibitem{huang2024stfdiff}
H.~Huang, W.~He, H.~Zhang, Y.~Xia, and L.~Zhang, ``Stfdiff: Remote sensing image spatiotemporal fusion with diffusion models,'' \emph{Information Fusion}, p. 102505, 2024.

\bibitem{huang2017arbitrary}
X.~Huang and S.~Belongie, ``Arbitrary style transfer in real-time with adaptive instance normalization,'' in \emph{Proceedings of the IEEE international conference on computer vision}, 2017, pp. 1501--1510.

\bibitem{HULLEY201957}
G.~C. Hulley, D.~Ghent, F.~M. Göttsche, P.~C. Guillevic, D.~J. Mildrexler, and C.~Coll, ``3 - land surface temperature,'' in \emph{Taking the Temperature of the Earth}.\hskip 1em plus 0.5em minus 0.4em\relax Elsevier, 2019, pp. 57--127.

\bibitem{li2013satellite}
Z.-L. Li \emph{et~al.}, ``Satellite-derived land surface temperature: Current status and perspectives,'' \emph{Remote sensing of environment}, vol. 131, pp. 14--37, 2013.

\bibitem{mhawej2022daily}
M.~Mhawej and Y.~Abunnasr, ``Daily ten-st-gee: An open access and fully automated 10-m lst downscaling system,'' \emph{Computers \& Geosciences}, vol. 168, p. 105220, 2022.

\bibitem{norman1995terminology}
J.~M. Norman and F.~Becker, ``Terminology in thermal infrared remote sensing of natural surfaces,'' \emph{Agricultural and Forest Meteorology}, vol.~77, no. 3-4, pp. 153--166, 1995.

\bibitem{roy2022anthropogenic}
P.~S. Roy \emph{et~al.}, ``Anthropogenic land use and land cover changes—a review on its environmental consequences and climate change,'' \emph{JISRS}, vol.~50, no.~8, pp. 1615--1640, 2022.

\bibitem{song2022mlff}
B.~Song, P.~Liu, J.~Li, L.~Wang, L.~Zhang, G.~He, L.~Chen, and J.~Liu, ``Mlff-gan: A multilevel feature fusion with gan for spatiotemporal remote sensing images,'' \emph{IEEE Transactions on Geoscience and Remote Sensing}, vol.~60, pp. 1--16, 2022.

\bibitem{zhan2016disaggregation}
W.~Zhan, F.~Huang, J.~Quan, X.~Zhu, L.~Gao, J.~Zhou, and W.~Ju, ``Disaggregation of remotely sensed land surface temperature: A new dynamic methodology,'' \emph{Journal of Geophysical Research: Atmospheres}, vol. 121, no.~18, pp. 10--538, 2016.

\bibitem{ZHANG2023129}
J.~Zhang and J.~Li, ``Chapter 11 - spacecraft,'' in \emph{Spatial Cognitive Engine Technology}, J.~Zhang and J.~Li, Eds.\hskip 1em plus 0.5em minus 0.4em\relax Academic Press, 2023, pp. 129--162.

\end{thebibliography}

\vspace{12pt}
\color{red}

\end{document}